\documentclass[twocolumn,10pt]{article}
\usepackage[T1]{fontenc}
\usepackage[utf8]{inputenc}
\usepackage{mathptmx} 
\usepackage[scaled=.90]{helvet} 
\usepackage{courier} 

\usepackage[a4paper,top=0.6in,bottom=0.6in,left=0.6in,right=0.6in]{geometry}
\usepackage{graphicx}
\usepackage{xcolor}
\usepackage{cuted}
\usepackage{microtype}
\usepackage{parskip}
\usepackage{tcolorbox}
\tcbuselibrary{skins,breakable}
\usepackage{academicons}
\usepackage{hyperref}
\usepackage{lipsum}
\usepackage{lastpage}
\usepackage{booktabs}
\usepackage{caption}
\usepackage{enumitem}
\usepackage{fancyhdr}
\usepackage{titlesec}
\usepackage{float}
\usepackage[ruled,vlined]{algorithm2e}
\usepackage{wrapfig}
\usepackage{amsmath, amssymb}
\usepackage{multicol}
\usepackage{amsthm}

\newtheorem{theorem}{Theorem}

\usepackage[backend=biber,style=ieee]{biblatex}
\definecolor{primary}{HTML}{003049}      
\definecolor{accent}{HTML}{F77F00}       
\definecolor{textgray}{HTML}{2F2F2F}
\definecolor{dividergray}{HTML}{DADADA}

\hypersetup{
  colorlinks=true,
  linkcolor=primary,
  citecolor=primary,
  urlcolor=primary
}

\titleformat{\section}
  {\color{primary}\sffamily\Large\bfseries}
  {\thesection}{1em}{}[\vspace{0.2em}\titlerule]

\titleformat{\subsection}
  {\color{primary}\sffamily\large\bfseries}
  {\thesubsection}{1em}{}

\titlespacing*{\section}{0pt}{6pt}{3pt}
\titlespacing*{\subsection}{0pt}{4pt}{2pt}

\setlist[itemize]{left=1.2em, itemsep=2pt, topsep=2pt, parsep=0pt}
\setlist[enumerate]{left=1.2em, itemsep=2pt, topsep=2pt, parsep=0pt}

\tcbset{
  abstractstyle/.style={
    enhanced,
    colback=white,
    colframe=primary,
    boxrule=1pt,
    fonttitle=\bfseries\sffamily\large,
    left=6mm, right=6mm, top=2mm, bottom=2mm,
    width=\textwidth,
    boxsep=4pt,
    breakable
  },
  biobox/.style={
    colback=white,
    colframe=primary,
    arc=1.5mm,
    boxrule=0.5pt,
    drop shadow southeast,
    left=3mm, right=3mm, top=1mm, bottom=1mm,
    width=\linewidth,
    before skip=6pt, after skip=6pt,
    breakable
  }
}

\SetAlgoNlRelativeSize{-1}
\SetAlCapNameFnt{\sffamily\bfseries\small}
\SetAlCapFnt{\sffamily\small}

\makeatletter
\def\@maketitle{%
  \begin{center}
    {\fontsize{26pt}{20pt}\selectfont \bfseries \textcolor{primary}{The Theory of the Unique Latent Pattern: A Formal Epistemic Framework for Structural Singularity in Complex Systems}}\\[1ex]
    {\normalsize 
\textbf{
Mohamed Aly Bouke\,%
\href{https://orcid.org/0000-0003-3264-601X}{\includegraphics[height=1.8ex]{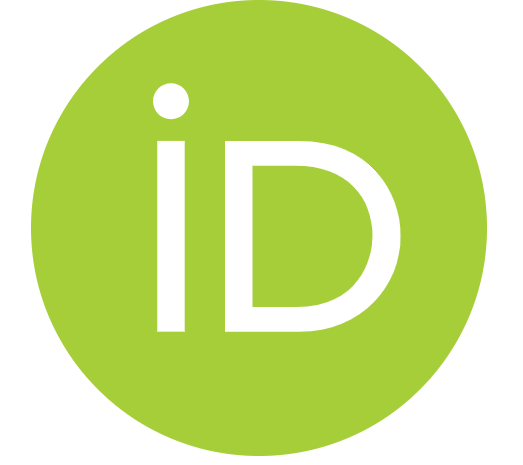}}%
\textsuperscript{1,*}
}
    }\\[0.8ex]
    {\footnotesize
      \textsuperscript{1}Department of Communication Technology and Network,\par Faculty of Computer Science and Information Technology, \par Universiti Putra Malaysia, Serdang 43400, Malaysia
    }\\[0.5ex]
    {\scriptsize \texttt{*bouke@ieee.org}}\\[1ex]
    {\scriptsize \textit{Short communication paper} — \today}
  \end{center}
}
\renewcommand{\maketitle}{%
  \twocolumn[
    \color{textgray}
    \@maketitle
    \vspace{-1.2em}
  ]
}
\makeatother

\begin{document}
\maketitle

\begin{strip}
  \begin{center}
    \begin{tcolorbox}[abstractstyle, title=Abstract]
      
      \normalsize
      This paper introduces the Theory of the Unique Latent Pattern (ULP), a formal epistemic framework that redefines the origin of apparent complexity in dynamic systems. Rather than attributing unpredictability to intrinsic randomness or emergent nonlinearity, ULP asserts that every analyzable system is governed by a structurally unique, deterministic generative mechanism, one that remains hidden not due to ontological indeterminacy, but due to epistemic constraints. The theory is formalized using a non-universal generative mapping \( \mathcal{F}_S(P_S, t) \), where each system \( S \) possesses its own latent structure \( P_S \), irreducible and non-replicable across systems. Observed irregularities are modeled as projections of this generative map through observer-limited interfaces, introducing epistemic noise \( \varepsilon_S(t) \) as a measure of incomplete access. By shifting the locus of uncertainty from the system to the observer, ULP reframes chaos as a context-relative failure of representation. We contrast this position with foundational paradigms in chaos theory, complexity science, and statistical learning. While they assume or model shared randomness or collective emergence, ULP maintains that every instance harbors a singular structural identity. Although conceptual, the theory satisfies the criterion of falsifiability in the Popperian sense, it invites empirical challenge by asserting that no two systems governed by distinct latent mechanisms will remain indistinguishable under sufficient resolution. This opens avenues for structurally individuated models in AI, behavioral inference, and epistemic diagnostics.

      \vspace{0.5em}

      \textbf{Keywords:} Structural uniqueness, Epistemic noise, Deterministic systems, Chaos reinterpretation, Popperian falsifiability
    \end{tcolorbox}
  \end{center}
\end{strip}

\section{Introduction} \label{sec:intro}
\vspace{0.8em}

The nature of complex systems has long puzzled scientists and philosophers alike. From the unpredictable trajectories of weather systems to the nonlinear fluctuations of financial markets and the individuality of human behavior, many real-world phenomena resist simple modeling or deterministic forecasting. In response to this resistance, scientific frameworks such as chaos theory and complexity science have emerged to make sense of the seemingly incoherent.

Chaos theory, for instance, highlights how systems governed by deterministic equations can nonetheless yield behaviors that appear random due to extreme sensitivity to initial conditions. Even minute changes in starting values can cause exponential divergence in outcomes, rendering long-term prediction practically impossible \parencite{lorenz1963deterministic}. Complexity science, on the other hand, focuses on how local interactions among many simple agents can produce emergent behaviors that are not easily reducible to their underlying components exemplified by everything from ant colonies to economies to neural networks \parencite{holland1992adaptation}.

While these frameworks have yielded profound insights, they also share a common philosophical stance that unpredictability is either intrinsic to the system or a byproduct of nonlinearity and scale. Under this view, there exists a kind of epistemic ceiling a limit beyond which no further clarity is attainable, at least not reliably.

\vspace{0.5em}

In this paper, we explore a competing perspective, the \textit{Theory of the Unique Latent Pattern} (ULP). Rather than viewing unpredictability as a fundamental property of certain systems, ULP Theory asserts that every system, no matter how complex or chaotic it appears follows a unique, internal, and structurally consistent pattern. The challenge lies not in the absence of such a pattern but in our inability to detect it due to limitations in our observational tools, analytical models, or cognitive frameworks.

This subtle shift from unpredictable systems to ``systems with undiscovered internal order has profound implications. It reframes chaos not as a natural boundary of knowledge, but as a signpost pointing to the temporary inadequacy of our current epistemic reach.

\section{Core Hypothesis}
\vspace{0.8em}
At the heart of the Theory of the ULP lies a single, provocative assertion:

\begin{quote}
\textit{Every system in the universe regardless of its scale, structure, or apparent disorder follows an internal pattern that is fixed, unique, and not replicated by any other system.}
\end{quote}

This hypothesis does not deny the presence of complexity, stochasticity, or sensitivity to initial conditions. Rather, it reframes these characteristics as surface-level manifestations of deeper, hidden structures. What appears chaotic or unpredictable is, in this view, merely a reflection of the observer’s epistemic limitations, not a property of the system itself.

\vspace{0.5em}

To ground the ULP within a rigorous formal structure, we define a system-theoretic framework that reflects both the ontological claims of the theory and its epistemic interpretation.

Let \( \mathcal{U} \) denote the space of all analyzable systems, understood as entities for which observable outputs can, in principle, be measured and modeled. Each system \( S \in \mathcal{U} \) is assumed to be governed by a latent internal configuration \( P_S \in \mathcal{P} \), where \( \mathcal{P} \) is the space of all structurally unique, system-specific generative patterns. These patterns encode not merely numerical parameters, but the underlying causal, structural, and formal rules that give rise to system behavior.

Let \( \mathbb{T} \) represent the temporal domain (discrete or continuous), and let \( \mathcal{O} \) represent the space of possible observable states. For each system \( S \), we define a generative map:

\begin{equation}
\mathcal{F}_S: \mathcal{P} \times \mathbb{T} \rightarrow \mathcal{O}
\end{equation}

which produces the idealized output \( O_S(t) \in \mathcal{O} \) at time \( t \in \mathbb{T} \), given the system's internal pattern \( P_S \). The central postulate of ULP is then formalized as:

\begin{equation}
\forall S \in \mathcal{U},\ \exists! \ P_S \in \mathcal{P} \ \text{such that} \ \mathcal{F}_S(P_S, t) = O_S(t)
\end{equation}

The uniqueness quantifier \( \exists! \) expresses the core claim of ULP: that each system is governed by a single latent configuration sufficient to generate its entire observable behavior over time, even when that behavior appears non-repeating or irregular.

In empirical settings, observers typically cannot access \( O_S(t) \) directly. Instead, observations are filtered through limitations imposed by the measurement apparatus, available context, and model abstraction. Let \( \tilde{O}_S(t) \) denote the recorded observation, and let \( \varepsilon_S(t) \in \mathcal{E} \) represent epistemic noise, where \( \mathcal{E} \) is the space of observer-dependent distortions including instrumental, contextual, and inferential uncertainties. Then:

\begin{equation}
\tilde{O}_S(t) = \mathcal{F}_S(P_S, t) + \varepsilon_S(t)
\end{equation}

This decomposition allows for several interpretive implications:

\begin{itemize}
  \item The map \( \mathcal{F}_S \) is system-specific. Each system has its own generative rule, which is not shared or generalized across systems. This reflects ULP's emphasis on structural non-replicability.

  \item The noise term \( \varepsilon_S(t) \) is epistemic, not ontological. It arises from the observer’s limitations, rather than being a feature of the system's internal dynamics.

  \item The equation is diagnostic rather than merely descriptive. It allows for inference about where a model fails, what remains hidden in \( \varepsilon_S(t) \) signals what is not yet captured in \( \mathcal{F}_S(P_S, t) \).

  \item The decomposition formalizes ULP as a theory of epistemic incompleteness, rather than ontological randomness. The unknown is not fundamentally unknowable, only not yet resolved.
\end{itemize}

Accordingly, unpredictability is not treated as a property of nature, but as a function of epistemic limitation. These limitations fall into three broad categories:

\begin{itemize}
  \item \textbf{Cognitive incompleteness}: lack of theoretical understanding of \( \mathcal{F}_S \) or \( P_S \).
  \item \textbf{Analytical insufficiency}: absence of tools or models capable of recovering or approximating \( P_S \) from data.
  \item \textbf{Observational constraint}: limited temporal or contextual coverage in empirical measurement.
\end{itemize}

In summary, this formal structure articulates ULP's central epistemological stance: that the apparent complexity or unpredictability of a system reflects the observer's incomplete grasp of its internal logic, not a failure of logic itself.

\subsection{Falsifiability}

The scientific legitimacy of the ULP does not rest on empirical accumulation alone, but on its alignment with the foundational criterion of falsifiability as articulated by Karl Popper \parencite{popper2005logic}. According to Popper, a theory is scientific not because it can be repeatedly confirmed, but because it exposes itself to potential refutation. In this view, the value of a theory lies not in its immunity to contradiction, but in the precision with which it invites it.

ULP makes a universal and bold epistemic claim, that every system no matter how complex, noisy, or chaotic in appearance, possesses a latent internal structure that is unique and, in principle, discoverable. This assertion is not a probabilistic generalization, but a categorical hypothesis about the nature of systems and the limits of knowledge. As such, it carries an inherent risk: a single well-constructed counterexample a deterministic system whose latent generator is provably irrecoverable, would falsify it. This vulnerability is not a flaw but a philosophical strength; it places ULP firmly within the domain of scientific discourse.

The Popperian strength of ULP lies in its formulation as a risky universal, a claim that applies to all systems, and thus is permanently exposed to decisive falsification. If epistemic limitation alone is the cause of perceived unpredictability, then the absence of pattern recovery cannot be attributed to the system itself. Rather, it must be attributed to the limitations of the observer. However, if it were shown that no conceivable method, regardless of depth or domain alignment, could in principle recover the latent structure of a fully deterministic system, then ULP would collapse under its own generality. This defines a clear, operational boundary for falsification.

Importantly, ULP does not require that such structure be currently observable with existing tools. It only maintains that unpredictability is not an ontological property of the system, but a contingent feature of our present epistemic apparatus. This distinction between in-principle discoverability and  practical accessibility  echoes Popper’s demarcation between truth and verifiability,  we may never fully verify the internal generator, but the theory's credibility grows with every attempt that fails to disprove it.

In this light, ULP embodies the Popperian ethos, it resists insulation from critique, refrains from metaphysical absolutes, and defines itself through its openness to disconfirmation. It is not a theory that seeks comfort in retrospective fitting, but one that courts contradiction by positing universal structure beneath the surface of chaos.

This Popperian grounding, while foundational, does not preclude engagement with alternative epistemological frameworks. From a Lakatosian viewpoint, the ULP may be interpreted as the “hard core” of a developing research programme—one whose falsifiability is not an isolated event but an evolving interaction between its theoretical nucleus and a protective belt of auxiliary models and measurement strategies \parencite{Lakatos_1978}. Such a framing shifts focus from isolated refutations to the program’s capacity for progressive theoretical refinement.

Conversely, Feyerabend’s epistemological anarchism challenges the privileging of falsifiability itself as a universal demarcation criterion \parencite{feyerabend1975against}. From this angle, even structurally unique hypotheses such as ULP should be assessed not by a fixed methodological rule, but by their heuristic power and capacity to generate novel conceptual and practical insights. By integrating such pluralist perspectives, the epistemic legitimacy of ULP becomes not a matter of strict formalism, but of its role in expanding the frontier of intelligibility in complex system analysis.

Thus, ULP’s scientific legitimacy is not merely methodological, it is philosophical. Its commitment to falsifiability is not an afterthought but an ontological stance: that the world is patterned, and that the limits of understanding lie not in nature itself, but in our evolving lens upon it.

\section{Components of the Theory}
\vspace{0.8em}

The ULP rests on three interdependent epistemic pillars. Each defines a necessary and theoretically rigorous condition for interpreting observable complexity as latent determinism constrained by observer knowledge, not system ontology.

\subsection{Structural Uniqueness}

ULP postulates that every analyzable system \( S \in \mathcal{U} \) is governed by a latent generator \( P_S \in \mathcal{P} \) that is structurally unique, that is, non-redundant and irreproducible across all other systems.

Formally, we define:

\[
\forall S_1, S_2 \in \mathcal{U},\ S_1 \neq S_2 \Rightarrow P_{S_1} \neq P_{S_2}
\]

Moreover, for their respective observable behaviors:

\[
\mathcal{F}_{S_1}(P_{S_1}, t) \not\equiv \mathcal{F}_{S_2}(P_{S_2}, t) \quad \text{for almost all } t \in \mathbb{T}
\]

This ensures that even if two systems exhibit superficially similar dynamics, they are generated by fundamentally distinct internal structures. Similarity is attributed to epistemic projection, not ontological identity.

This principle rules out the possibility of shared archetypes or statistical averaging of behavior, placing ULP in sharp contrast with both latent variable models and complexity frameworks that permit overlapping generative dynamics.

\subsection{Conditional Discoverability}

While latent patterns \( P_S \) may be obscured in practice, ULP asserts that they are recoverable in principle, given sufficient epistemic access. Discoverability is thus a function of the observer’s cognitive, computational, and contextual resolution.

Let \( \mathcal{R} \) be an epistemic reconstruction operator such that:

\[
\mathcal{R}: \tilde{O}_S(t) \rightarrow \hat{P}_S
\]

We define conditional discoverability as:

\[
\exists\ \mathcal{R}_\theta\ \text{such that}\ \| \hat{P}_S - P_S \| < \varepsilon, \quad \text{given}\ \theta \in \Theta
\]

Where:\\
- \( \hat{P}_S \) is the reconstructed pattern under the observer’s model class.\\
- \( \varepsilon \) is an acceptable approximation threshold\\
- \( \Theta \) is the epistemic condition space comprising:\\
  - domain knowledge\\
  - analytical tooling\\
  - observational breadth.\\

This defines ULP not as a theory of omniscience, but of asymptotic recoverability, where patterns are knowable not instantly, but increasingly, as epistemic reach expands.

\subsection{Epistemic Reframing of Chaos}

Conventional interpretations view chaos as a signature of ontological indeterminism, systems that are inherently unpredictable. ULP reframes this view epistemically, unpredictability does not reflect a property of the system, but a boundary condition of the observer.

Let \( \mathcal{E} \) be the epistemic error space (instrumental, inferential, contextual). Then:

\[
\tilde{O}_S(t) = \mathcal{F}_S(P_S, t) + \varepsilon_S(t),\quad \varepsilon_S(t) \in \mathcal{E}
\]

If:

\[
\lim_{\mathcal{E} \to 0} \| \mathcal{R}_\theta(\tilde{O}_S) - P_S \| = 0
\]

then unpredictability is epistemic, not intrinsic. Conversely, if this limit fails to converge even under theoretically maximal access, ULP is falsified.

This logic adheres to Popperian falsifiability, ULP makes a strong, refutable claim namely, that what appears chaotic today is the result of unmodeled regularity, not fundamental randomness.

Thus, ULP converts “chaos” into a moving frontier, an ever-shifting boundary between what is known and what can, in principle, be known.

\section{Relationship to Existing Theories}
\vspace{0.8em}

To situate ULP meaningfully, it is necessary to examine where it aligns with existing theories, and more importantly, where it diverges from them in formal, ontological, and predictive structure.

Classical chaos theory, for instance, operates on deterministic systems of the form:
\begin{equation}
x_{n+1} = f(x_n), \quad f \in \mathcal{C}^1
\end{equation}
where the unpredictability arises from sensitivity to initial conditions. The map \( f \) is typically held constant across the system class, variability is attributed to \( x_0 \). In contrast, ULP posits that unpredictability does not emerge from within a shared function space, but rather from the observer's lack of access to a unique, system-specific map \( \mathcal{F}_S \). More formally:
\begin{equation}
\forall S \in \mathcal{U},\ \exists! \ \mathcal{F}_S \text{ such that } O_S(t) = \mathcal{F}_S(P_S, t)
\end{equation}
Here, each system is governed not by a universal function, but by its own structurally irreducible rule, a generative identity, not a parameter variant. Where chaos theory decentralizes individuality in favor of collective dynamics, ULP reintroduces singularity at the ontological core.

This divergence becomes more explicit when contrasted with complexity science, which seeks to explain global phenomena through micro-level interactions. A typical formulation might involve:
\begin{equation}
\vec{x}_{t+1} = \phi(\vec{x}_t, \vec{N}_t)
\end{equation}
where \( \vec{N}_t \) denotes the neighborhood function and \( \phi \) defines interaction rules. Such models do not assume pre-existing structure but allow structure to "emerge" from interaction. ULP, however, challenges the sufficiency of emergence as an explanatory principle. It argues that even emergent phenomena presuppose internal constraints that are not accounted for by relational dynamics alone. The latent generator \( P_S \) in ULP is not the product of emergence; it is the silent architect behind what appears emergent.

From a statistical modeling standpoint, latent variable frameworks such as factor analysis, variational autoencoders, and state-space models rely on the assumption of shared generative distributions. In these frameworks, data \( X \in \mathbb{R}^n \) is modeled as:
\begin{equation}
X = g(Z) + \varepsilon
\end{equation}
where \( Z \sim p(Z) \) and \( g \) is a universal or learned mapping. This assumption entails that all observations, though distinct, are generated from the same latent space \( p(Z) \), implying a form of compressibility across instances. ULP refutes this. It holds that for each data generating entity \( S \), the latent generator \( P_S \) is unique, and that the function \( \mathcal{F}_S \) mapping \( P_S \) to observable trajectories is non-transferable. There exists no global \( g \), only an ensemble of singular mappings:
\begin{equation}
x_t^{(i)} = \mathcal{F}_{S_i}(P_{S_i}, t), \quad \text{with } \mathcal{F}_{S_i} \ne \mathcal{F}_{S_j} \ \text{for } i \ne j
\end{equation}
Thus, ULP renders generalization over instances insufficient and advocates for differentiation at the modeling level itself.

Even philosophical traditions such as constructivist epistemology, where knowledge is treated as a mental construction shaped by context, are extended, not rejected, by ULP. Constructivism contends that all perception is filtered through interpretive frameworks, but often limits this insight to the observer. ULP expands it to the system, it suggests that each system possesses not only an epistemic uniqueness (in how it is seen) but a structural one (in how it functions), regardless of whether it is seen at all. This transition from epistemic relativism to structural realism is subtle yet profound, it implies that hidden patterns are not only discoverable but ontologically embedded.

Beyond formal comparisons with dynamical and statistical frameworks, ULP also invites interpretation through the lens of contemporary philosophy of science. Notably, Nancy Cartwright’s pragmatic realism asserts that scientific models “lie” in the sense that they work locally without representing global truths \parencite{cartwright1983laws}. While ULP shares this sensitivity to context, it radicalizes the stance by proposing that each system instantiates its own structurally unique generator, irreducible even within local clusters of behavior. 

Similarly, Ian Hacking’s view that scientific models both represent and intervene \parencite{hacking1983representing} resonates with ULP’s emphasis on epistemic noise as a function of theoretical framing. Yet ULP reframes this interventionism diagnostically: the distortion in observation is not only a byproduct of experimental act, but a signal of what remains unresolved in the observer’s conceptual scope.

These philosophical alignments suggest that ULP does not merely propose a new modeling strategy, but offers a metaphysical commitment to structural singularity grounded in epistemic humility. It thus extends the epistemological discourse beyond anti-realism and instrumentalism toward a differentiated realism anchored in irreducibility.

Ultimately, ULP's most radical claim is not that patterns exist, but that no two systems share the same one. This disrupts the statistical ethos of modern science, which privileges commonality, and reasserts singularity as a valid scientific object. Where most models seek what systems have in common, ULP demands attention to what makes each system unrepeatable. This departure transforms not only our formal tools but the very philosophical stance from which modeling proceeds.

\section{Proof of Concept}
\vspace{0.8em}

While the ULP is rooted in epistemic and ontological principles, it admits a class of indirect conceptual proofs through mathematical modeling of uniqueness and separability under well-defined constraints. Rather than constructing empirical experiments, we outline a general strategy for how uniqueness can be deduced under the framework of functional distinguishability in dynamical systems.

Let us consider a set of deterministic systems \( \{S_i\}_{i=1}^N \), each governed by a distinct latent structure \( P_{S_i} \in \mathcal{P} \), and let \( \mathcal{F}_{S_i} \) denote the internal generative mechanism that maps latent structure and time into the observable domain:
\begin{equation}
x_t^{(i)} = \mathcal{F}_{S_i}(P_{S_i}, t)
\label{eq:ulp-system}
\end{equation}

The central question is, under what conditions are the mappings \( \mathcal{F}_{S_i} \) functionally distinguishable in the observation space \( \mathcal{O} \)? That is, does:
\begin{equation}
\mathcal{F}_{S_i}(P_{S_i}, t) \neq \mathcal{F}_{S_j}(P_{S_j}, t) \quad \text{for all } i \neq j
\label{eq:distinct-maps}
\end{equation}
imply observable separability even when the outputs \( x_t^{(i)} \) and \( x_t^{(j)} \) may overlap in a coarse representation (e.g., due to measurement compression or projection)?

Let \( \Phi: \mathcal{O}^T \rightarrow \mathbb{R}^k \) be an admissible transformation (e.g., a projection, integral operator, or dimension-reduction map) satisfying:
\begin{equation}
\text{If } \mathcal{F}_{S_i} \not\equiv \mathcal{F}_{S_j}, \text{ then } \Phi(x_t^{(i)}) \neq \Phi(x_t^{(j)})
\label{eq:injectivity}
\end{equation}

This injectivity condition over transformed trajectories provides a theoretical criterion for the recoverability of system-specific identity from observable data. The existence of such transformations (e.g., spectral decompositions, topological embeddings, or symbolic encodings) is guaranteed in classes of functions that are Lipschitz continuous, bounded, and structurally non-redundant assumptions well-supported in the study of nonlinear dynamical systems and information geometry.

Thus, the “proof of concept” in the ULP context is not empirical in the conventional sense, but epistemic, if two systems produce observables generated from non-equivalent mappings, then under sufficient analytical resolution, their outputs will be functionally separable. This leads to a general theorem schema:

\begin{theorem}
Let \( S_i, S_j \in \mathcal{U} \) be deterministic systems such that \( \mathcal{F}_{S_i} \not\equiv \mathcal{F}_{S_j} \). Then there exists a transformation \( \Phi \) such that:
\[
\Phi\left(\mathcal{F}_{S_i}(P_{S_i}, t)\right) \neq \Phi\left(\mathcal{F}_{S_j}(P_{S_j}, t)\right)
\]
for all \( t \in \mathbb{T} \), provided \( \mathcal{F}_{S_i} \) and \( \mathcal{F}_{S_j} \) belong to a class of structurally expressive maps (e.g., analytic or piecewise continuous).
\end{theorem}

This formulation aligns with ULP’s philosophical foundation, the uniqueness of latent structure implies, in principle, observable distinction contingent not on the complexity of the system, but on the epistemic power of the analytical frame. What appears as unpredictability is merely an artifact of projection, not a collapse of determinism.

Accordingly, this section reframes the notion of validation in ULP, not as statistical confirmation from simulations, but as logical and mathematical consistency within a general class of functionally unique systems.

\section{Potential Applications}
\vspace{0.8em}

The ULP offers more than a philosophical reframing of chaos and complexity, it provides a practical lens through which diverse applied domains may be reconceptualized. By interpreting unpredictability not as an intrinsic feature of systems but as a reflection of epistemic limitation, ULP shifts the focus of model design from generalization across instances to the discovery of internal structure within each one.

This paradigm shift has important implications for modeling human behavior. Whereas traditional behavioral science often relies on heuristic or probabilistic representations, grouping individuals into statistical archetypes, ULP suggests that each person may operate according to a distinct, internally consistent behavioral pattern. With sufficient high-resolution data (such as keystroke dynamics, interaction histories, or psychometric timelines) and advanced inference techniques like symbolic AI or neural representation learning, it becomes possible to approximate an individualized function \( f_{\text{person}}(s, t) \). This approach opens pathways for personalized mental health diagnostics, adaptive user interfaces, and deeper models of human–machine interaction \parencite{costa1999five}.

A similar reconceptualization applies to economic forecasting. Conventional macroeconomic and market models aggregate behavior across agents, often sacrificing individual specificity for tractability. In contrast, ULP suggests that each market, or even each actor within a market, follows a latent decision engine that is unique to its structure and context. Techniques such as symbolic regression, agent-based reinforcement learning, and structural causal modeling can be used to uncover these hidden engines \parencite{koza1992evolution,sutton1998reinforcement}. Rather than forecasting based on surface-level correlations, ULP, inspired models would infer deep behavioral rules tailored to each economic subsystem. This could enable new forms of microeconomic modeling, sovereign risk analysis, and resilient supply chain design.

In the realm of education, ULP introduces a compelling direction for adaptive learning. Most current platforms adjust difficulty using surface metrics like test scores or response time, yet these often fail to capture how a learner conceptualizes or internalizes knowledge. ULP instead motivates the search for a latent cognitive function \( f_{\text{student}}(x, t) \) that governs how a specific learner thinks, evolves, and interacts with information. Sequence models and attention-based architectures \parencite{vaswani2017attention} can be designed not just to respond to correct or incorrect answers, but to trace the learner’s conceptual trajectory. Such models could shift the paradigm from reactive calibration to proactive personalization.

Lastly, ULP proposes a novel trajectory for the development of artificial intelligence itself. Traditional machine learning emphasizes generalization, training models to find shared patterns across data. By contrast, ULP-inspired exploratory AI would prioritize differentiation, seeking to identify what is uniquely true about each sample or system. Instead of learning a single global function \( f(x) \), such systems would aim to discover local functions \( f_i(x) \) that reflect the inner dynamics of each individual instance. This methodology could support applications such as one-shot personalization, precision modeling in synthetic biology, or anomaly detection in cybersecurity, where singularity, not typicality, carries the most significance.

Across these domains, ULP reorients applied modeling away from statistical averages and toward structural individuality. Its practical value lies not only in enhancing prediction, but in refining attention. It compels us to reconsider what we often overlook, that uniqueness is not deviation or noise, but signal, a structural feature worthy of direct inquiry.

\section{Conclusion}
\vspace{0.8em}

The ULP advances a principled reconceptualization of complexity, reframing unpredictability not as a feature of ontological randomness, but as an artifact of epistemic limitation. It proposes that beneath every observable behavior, however erratic or nonlinear it may appear, lies a structurally unique and internally deterministic generator, singular to that system and non-replicable in any other.

This postulate disrupts the classical dichotomy between determinism and stochasticity by locating the boundary of unpredictability not within the system itself, but in the representational constraints of the observer. Chaos, in this framing, is neither noise nor unknowability, it is a diagnostic category a signal of analytical insufficiency rather than ontological indeterminacy.

Mathematically, ULP embeds this claim in a formal structure, each system \( S \in \mathcal{U} \) possesses a latent configuration \( P_S \) and a system-specific generative map \( \mathcal{F}_S \), such that the observable trajectory \( O_S(t) \) is uniquely determined by \( \mathcal{F}_S(P_S, t) \). When coarse-grained into the observed form \( \tilde{O}_S(t) \), the divergence between theory and data becomes attributable not to randomness, but to a noise term \( \varepsilon_S(t) \) that encodes epistemic deficits in instrumentation, modeling, or context.

From a philosophy of science perspective, ULP adheres to the Popperian criterion of falsifiability. Its universal claim, that every system is uniquely structured exposes it to risk, a single demonstrable counterexample would refute it. Yet this vulnerability is its strength. The theory does not seek immunity through generality, but invites precise empirical challenges through structural distinctness. It reframes science not as the discovery of universal laws across systems, but as the recovery of irreducible rules within each.

Rather than appealing to simulation or induction alone, ULP emphasizes the recoverability of latent identity through transformations that preserve uniqueness under constraints of resolution and context. This orientation opens space for a new class of analytical methods, ones that prioritize differentiation over aggregation, structure over statistics, and identity over generality.

\section*{Declarations}
\vspace{0.8em}
\begin{itemize}
  \item \textbf{Funding:} Not applicable.
  
  \item \textbf{Conflict of Interest:} The authors declare that there is no conflict of interest.

\item \textbf{Availability of Data and Materials: } 
This study did not use any  datasets.

  \item \textbf{Ethics Approval:} Not applicable.
\end{itemize}

\section*{Author Biographies}
\vspace{0.8em}

\begin{tcolorbox}[biobox]
  \begin{wrapfigure}{l}{0.35\linewidth}
    \vspace{-0.5em}
    \includegraphics[width=\linewidth]{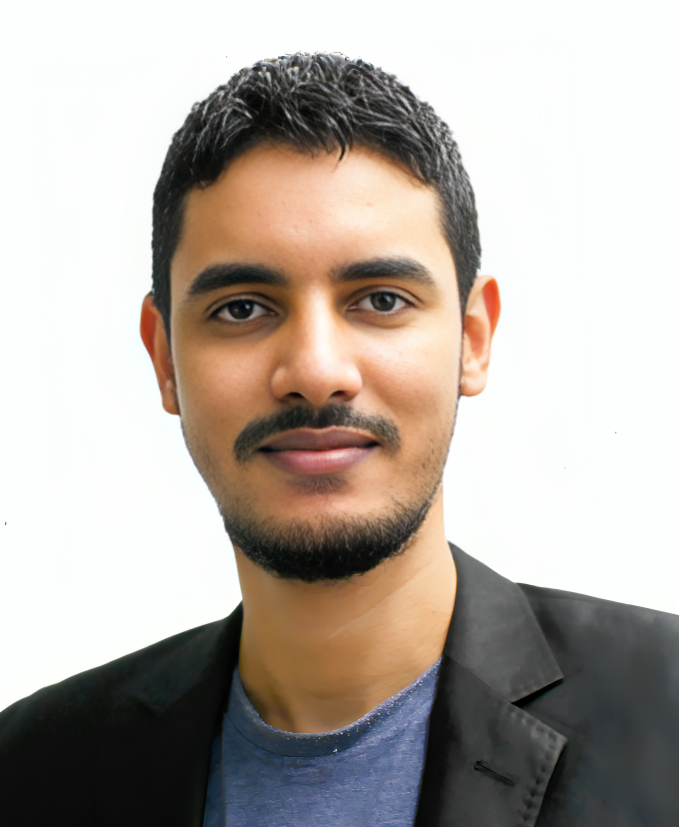}
  \end{wrapfigure}
  \textbf{Mohamed Aly Bouke}\,\href{https://orcid.org/0000-0003-3264-601X}{\includegraphics[height=1.8ex]{orcid.png}} is a researcher with interdisciplinary expertise across theoretical mathematics, computer science, artificial intelligence, and cryptography. He holds a Master’s and a Ph.D. in Information Security from Universiti Putra Malaysia and has a background in mathematics education. His academic work spans topics such as mathematical modeling, epistemic systems, AI architectures, and secure computation. Dr. Bouke is an active member of the \textit{Institute of Electrical and Electronics Engineers (IEEE)}, the \textit{International Information System Security Certification Consortium (ISC2)}, and the \textit{Institute for Systems and Technologies of Information, Control and Communication (INSTICC)}. His contributions include peer-reviewed publications, invited talks, and academic training in both technical and theoretical domains.

  \vspace{0.8em}
  \textit{Email: bouke@ieee.org}
\end{tcolorbox}

\printbibliography

\end{document}